\documentclass[10pt,twocolumn,letterpaper]{article}

\usepackage{iccv}
\usepackage{times}
\usepackage{epsfig}
\usepackage{graphicx}
\usepackage{amsmath}
\usepackage{amssymb}

\usepackage{booktabs}
\usepackage{multirow}
\usepackage{marvosym}

\usepackage[breaklinks=true,bookmarks=false]{hyperref}

\iccvfinalcopy 

\def\iccvPaperID{****} 

\ificcvfinal\fi

\begin{document}

\title{ClearSight: Human Vision-Inspired Solutions for Event-Based \\ Motion Deblurring}

\author{Xiaopeng Lin\textsuperscript{*}, Yulong Huang\textsuperscript{*}, Hongwei, Ren, Zunchang Liu, Yue Zhou, Haotian Fu, Bojun Cheng\Letter\\
The Hong Kong University of Science and Technology (Guangzhou)\\
{\tt\small \{xlin746, yhuang496, hren066, zliu361, yzhou883, hf373\}@connect.hkust-gz.edu.cn} \\
{\tt\small bocheng@hkust-gz.edu.cn}
}

\maketitle
\ificcvfinal\thispagestyle{empty}\fi

\begin{abstract}

Motion deblurring addresses the challenge of image blur caused by camera or scene movement. Event cameras provide motion information that is encoded in the asynchronous event streams. To efficiently leverage the temporal information of event streams, we employ Spiking Neural Networks (SNNs) for motion feature extraction and Artificial Neural Networks (ANNs) for color information processing. Due to the non-uniform distribution and inherent redundancy of event data, existing cross-modal feature fusion methods exhibit certain limitations. Inspired by the visual attention mechanism in the human visual system, this study introduces a bioinspired dual-drive hybrid network (BDHNet). Specifically, the Neuron Configurator Module (NCM) is designed to dynamically adjusts neuron configurations based on cross-modal features, thereby focusing the spikes in blurry regions and adapting to varying blurry scenarios dynamically. Additionally, the Region of Blurry Attention Module (RBAM) is introduced to generate a blurry mask in an unsupervised manner, effectively extracting motion clues from the event features and guiding more accurate cross-modal feature fusion. Extensive subjective and objective evaluations demonstrate that our method outperforms current state-of-the-art methods on both synthetic and real-world datasets.
\end{abstract}

\section{Introduction}

Motion blurring primarily occurs due to the movement of either the camera or the moving objects during the sensor's exposure period \cite{li2023real, liadaptive}. Deblurring is a critical task focused on recovering a sharp image with clear details from the motion-blurred counterpart. Several image-based deblurring approaches have been developed to compensate for the blur characteristics with enhanced performance, including traditional approaches\cite{fergus2006removing, bahat2017non} and learning-based approaches \cite{cho2021rethinking, tsai2022stripformer, chen2021hinet}. However, deblurring methods that rely solely on conventional frame-based cameras frequently face performance constraints due to the absence of essential motion information. These limitations are particularly pronounced under adverse lighting conditions and during the capture of rapidly moving objects.

\begin{figure}[t]
\begin{center}

\includegraphics[width=1\linewidth]{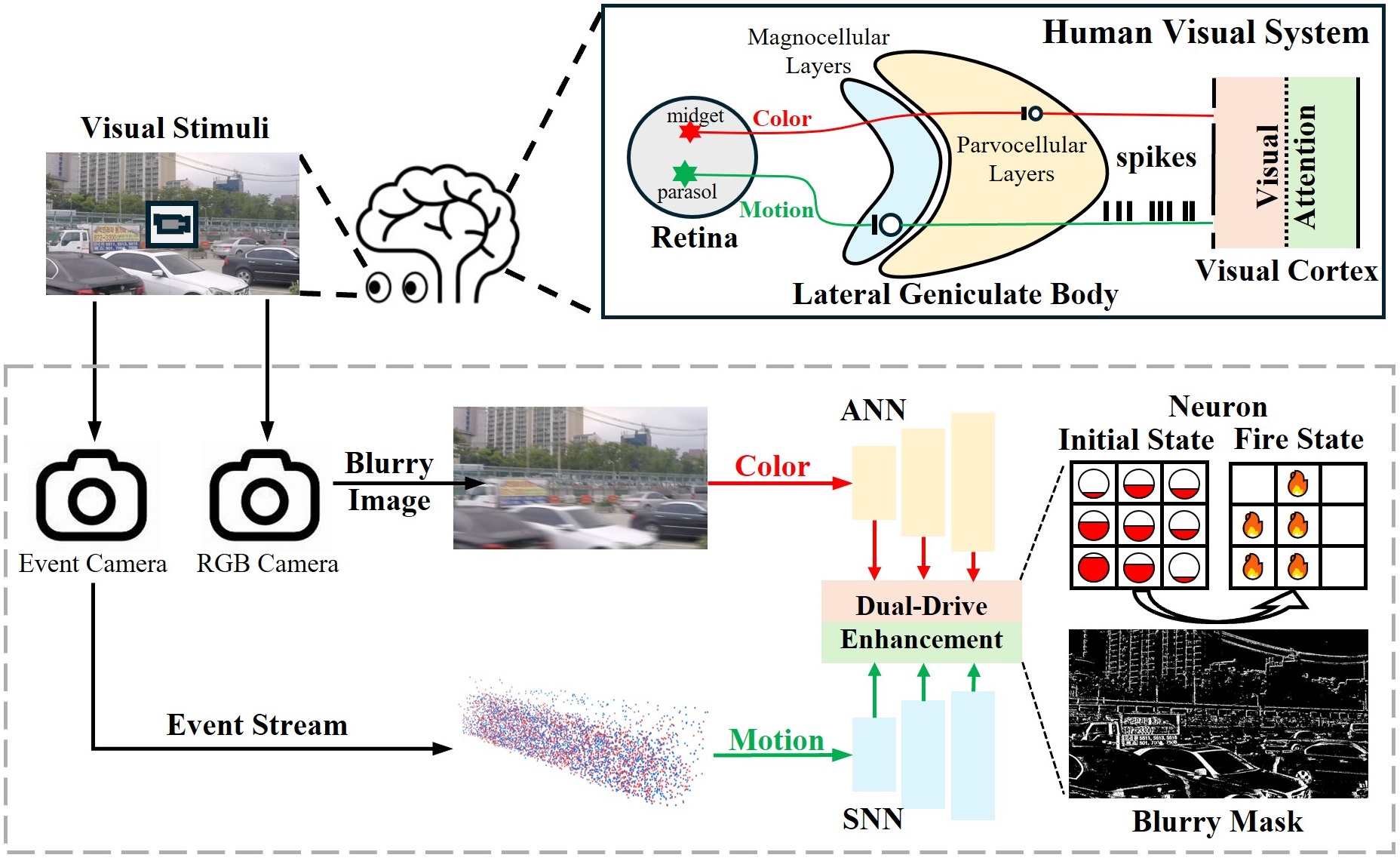}
   \caption{The working mechanism of human visual system after receiving visual stimuli and the proposed bio-inspired dual-drive hybrid network. The visual attention in visual cortex consists of the neuron-based (Pink) and synapse-based attention (Green) \cite{kanwisher2000visual}. More details are provided in the supplementary material.}
\label{fig:motivation}
\end{center}
\end{figure}

Drawing inspiration from biological systems, event-based cameras introduce an innovative paradigm for visual data acquisition \cite{brandli2014240, gallego2020event}. Event camera captures changes in brightness in high temporal resolution that naturally emphasize high-contrast edges. The explicit contrast edge information and the implicit temporal correlation among the event streams help to recover details lost in blurry images \cite{zhang2022unifying, lin2024event}. Effective integration of event streams and frame-based images requires the precise extraction of motion features from event streams and background details from blurry images. However, mainstream ANN methods integrate event streams into frame-based or voxel-based representations and process in different channels, losing the temporal dependency \cite{liu2023motion}. 

Spiking Neural Networks (SNNs)\cite{bouvier2019spiking} are inherently suitable to handle asynchronous event streams. Visual systems integrating SNNs with event cameras demonstrate promising performance in addressing complex visual tasks, attributed to SNNs' capability to effectively preserve the temporal dependencies inherent in event data \cite{ren2023spikepoint, cao2023event, fu2024event, lin2024event}. Nevertheless, image restoration is distinct from other high-level tasks that cannot be entirely based on SNNs for feature extraction. This is because SNNs encode information via binary spike sequences \cite{teeter2018generalized}, which are insufficient for preserving essential high bit details such as color and structure information. This limitation affects the pixel-level precision required for image restoration. For this reason, hybrid networks with Artificial Neural Networks (ANNs) for color information processing and Spiking Neural Networks (SNNs) for event processing are adopted, aiming to combine the advantages of both ANN and SNN \cite{fu2024event, cao2023event, liu2023motion}. 

To effectively integrate data from both sensor types, previous studies \cite{sun2022event, chen2024motion, yang2024motion} have concentrated on the development of cross-modal attention mechanisms, achieving substantial performance improvements. However, previous synaptic-level attention mechanisms that modify weight magnitudes overlook the inherent physical differences between sensor modalities, limiting the effectiveness. This limitation arises from two main factors: Firstly, the distribution of events is non-uniform. In the blurry regions, events generated by low-contrast scenes exhibit a sparse distribution. This sparsity is insufficient to trigger neuron responses in SNNs, resulting in the ineffectiveness of cross-modal attention mechanisms. Second, event features exhibit redundancy. Within the exposure duration, numerous events are triggered by repetitive movements associated with the same objects or the contours that are the details typically lost in blurry images. Existing approaches that utilize all event features for cross-modal attention fail to effectively discriminate the specific motion features responsible for the blurriness in these areas, leading to suboptimal performance in motion deblurring.  

To better leverage the multi-modal information for Event-based Motion Deblurring, this paper introduces a Bioinspired Dual-Drive Hybrid Network (BDHNet). Specifically, we designed a Neuron Configurator Module (NCM) as visual enhancement from image data to event data, to improve the performance in blurry regions with sparse events. The NCM module utilizes image features for cross-modal initialization of the neurons’ membrane potential and threshold in the SNN blocks, enabling a pixel-level dynamic adjustment. Such behavior is categorized as neuron-based attention, which is also present in the human visual system as the baseline increase in neural activity that elevates neuron activity across specific visual areas \cite{kanwisher2000visual}. The configuration enables more neural responses in the blurry regions when the event stream is sparse, and significantly enhances the SNN’s capacity to extract detailed motion features. For the precise motion clue extraction, we introduce a Region of Blurry Attention module (RBAM) to enhance the synapse-based attention as visual enhancement from event data to image data. The RBAM module integrates localized spike data with image features to generate a mask specifically targeting blurry regions. This mask is subsequently utilized to selectively recalibrate the cross-modal feature fusion, strategically concentrating the network’s perceptual focus on blurry areas, thereby enhancing the deblurring performance. The main contributions of our work are summarized as follows:  

\begin{itemize}

\item We propose a bioinspired dual-drive hybrid network with the neuron-based and enhanced synapse-based attention to mimic the visual attention capability of the human visual system.  

\item We introduce a Neuron Configurator Module for the dynamic configurations of the SNN neurons and a Region of Blurry Attention Module that creates a targeted blurry mask to facilitate cross-modal feature fusion.

\item Subjective and objective evaluations demonstrate that our BDHNet has achieved SOTA in varying blurry conditions in GoPro, REBlur and MS-RBD datasets.

\end{itemize}

\section{Background and Related Work}


\subsection{Event-based Motion Deblurring}

Event cameras capture continuous motion data with low latency, providing vital cues for enhancing motion deblurring. Recent researches have achieved notable advancements and demonstrate the effectiveness of integrating event data. EDI \cite{pan2019bringing} establishes a rigorous mathematical integration between blurry images, event data, and sharp reference frames. eSL-Net \cite{wang2020event} applies sparse learning to simultaneously denoise and enhance the resolution of images shaped from event data, effectively restoring high-quality results. EVDI \cite{zhang2022unifying} presents a comprehensive framework for event-based motion deblurring and frame interpolation, utilizing the low latency of event cameras to mitigate motion blur and enhance frame prediction. DS-Deblur \cite{yang2022learning} implements a dual-stream architecture that combines adaptive feature fusion with recurrent spatio-temporal transformations, refining image clarity. GEM \cite{zhang2023generalizing} introduces a scale-aware network that adjusts to varying spatial and temporal scales, employing a self-supervised learning strategy to adapt to diverse real-world scenarios. MTGNet \cite{lin2024event} proposes multi-temporal granularity network that efficiently merges voxel-based and point cloud-based events to optimize the exploitation of the inherent high temporal resolution.

Recent advancements leverage cross-modal attention mechanisms for effective multi-modal integration, achieving notable enhancements. EFNet \cite{sun2022event} incorporates a multi-head attention mechanism to integrate data across different modalities. EIFNet \cite{yang2023event} improves motion deblurring by efficiently processing both unique and shared features through a dual cross-attention mechanism, enhancing feature integration and differentiation. MAENet \cite{sun2025motion} utilizes alignment and multi-head attention to coherently fuse features, reducing inter-modal inconsistencies. STCNet \cite{yang2024motion} implements differential-modality calibration and co-attention to enhance spatial fusion and model cross-temporal dependencies between frames and events using motion information. 

Despite significant progress in event-based image deblurring, current methodologies exhibit fundamental limitations. Firstly, attention mechanisms at the synaptic level that only adjust the weight magnitudes are insufficient limited to the non-uniform distribution and inherent redundancy of event data. Moreover, conventional CNN-based approaches fail to adequately preserve the intrinsic temporal dependencies of event data, reducing the overall effectiveness of the deblurring process.

\subsection{Spiking Neural Networks}

Spiking Neural Networks, as bioinspired computational frameworks, are inherently suited to handle the asynchronous and sparse characteristics of event data \cite{fang2021deep, huangclif}. The most common neuron model is the Leaky Integrate-and-Fire (LIF) model with iterative expression \cite{wu2018spatio}. At each timestep $t$, the neurons in the $l$-th layer integrate the postsynaptic current $\boldsymbol{c}^l[t]$ with previous membrane potential $\boldsymbol{u}^{l}[t-1]$, the mathematic expression is illustrated in Equation~\eqref{eq:preliminary:ut = ut-1}:
\begin{equation}
\boldsymbol{u}^{l}[t] = (1 - \frac{1}{\tau}) \boldsymbol{u}^l[t - 1]  + \boldsymbol{c}^l[t], 
\label{eq:preliminary:ut = ut-1}
\end{equation}
where $\tau$ is the membrane time constant. $\tau > 1$ as the discrete step size is 1. The postsynaptic current $\boldsymbol{c}^l[t] = \mathcal{W}^l * \boldsymbol{s}^{l-1}[t]$ is calculated as the product of weights $\mathcal{W}^l$ and spikes from the preceding layer $\boldsymbol{s}^{l-1}[t]$, simulating synaptic functionality, with $*$ indicating either a fully connected or convolutional synaptic operation.

Neurons produce spikes $\boldsymbol{s}^l[t]$ via the Heaviside function $\Theta$ when the membrane potential $\boldsymbol{u}^l[t]$ surpasses the threshold $V_{\mathrm{th}}$, as depicted in Equation~\eqref{eq:preliminary:s=f(u-vth)}:
\begin{equation}
\boldsymbol{s}^l[t]  = \Theta (\boldsymbol{u}^{l}[t] - V_{\mathrm{th}}) = \begin{cases}
1, & \text{if }  \boldsymbol{u}^l[t] \geq V_{\mathrm{th}} \\
0, & \text{otherwise}
\label{eq:preliminary:s=f(u-vth)}
\end{cases} .
\end{equation}
After the spike, the neuron updates the membrane potential $\boldsymbol{u}^l[t]$ according to the reset mechanism as shown in Equation~\eqref{eq:preliminary:reset}:
\begin{equation}
\boldsymbol{u}^{l}[t] =
 \boldsymbol{u}^l[t] - V_{\mathrm{th}} \boldsymbol{s}^l[t]
\label{eq:preliminary:reset},
\end{equation}
where the $V_{\mathrm{th}} \in \mathbb{R}$ is generally a global scalar that controls the firing and reset process for the neurons in each layers.

\subsubsection{SNN-based Image Restoration}

Recent works leverage SNNs for effective multi-modal image restoration, achieving impressive results. EMFHNet \cite{li2022image} introduces an event-enhanced multi-modal fusion hybrid network, incorporating an SNN encoder to efficiently process and denoise event data. SC-Net \cite{cao2023event} effectively combines SNNs and CNNs to exploit the sparse temporal and spatial characteristics of event stream, enhancing event-driven video restoration. ESDNet \cite{song2024learning} designs spiking residual block and attention mechanisms to enhance image deraining, effectively addressing the challenges of binary activation and complex training dynamics. EDHNet \cite{fu2024event} introduces a hybrid event-driven network with a bimodal fusion module to effectively identify and remove rain streaks, significantly improving video deraining. MotionSNN \cite{liu2023motion} employs a spiking neural network and a hybrid feature extraction encoder to optimize event-based image deblurring, seamlessly merging high-temporal-resolution event data with the image data for enhanced clarity.

However, current spike-based image restoration methods, with the uniform neuron configurations in the SNN branch, lack adaptability to the non-uniform distribution of event data and fail to harness complementary multi-modal inputs, resulting in compromised performance.

\begin{figure*}[ht]
\begin{center}
\includegraphics[width=1\linewidth]{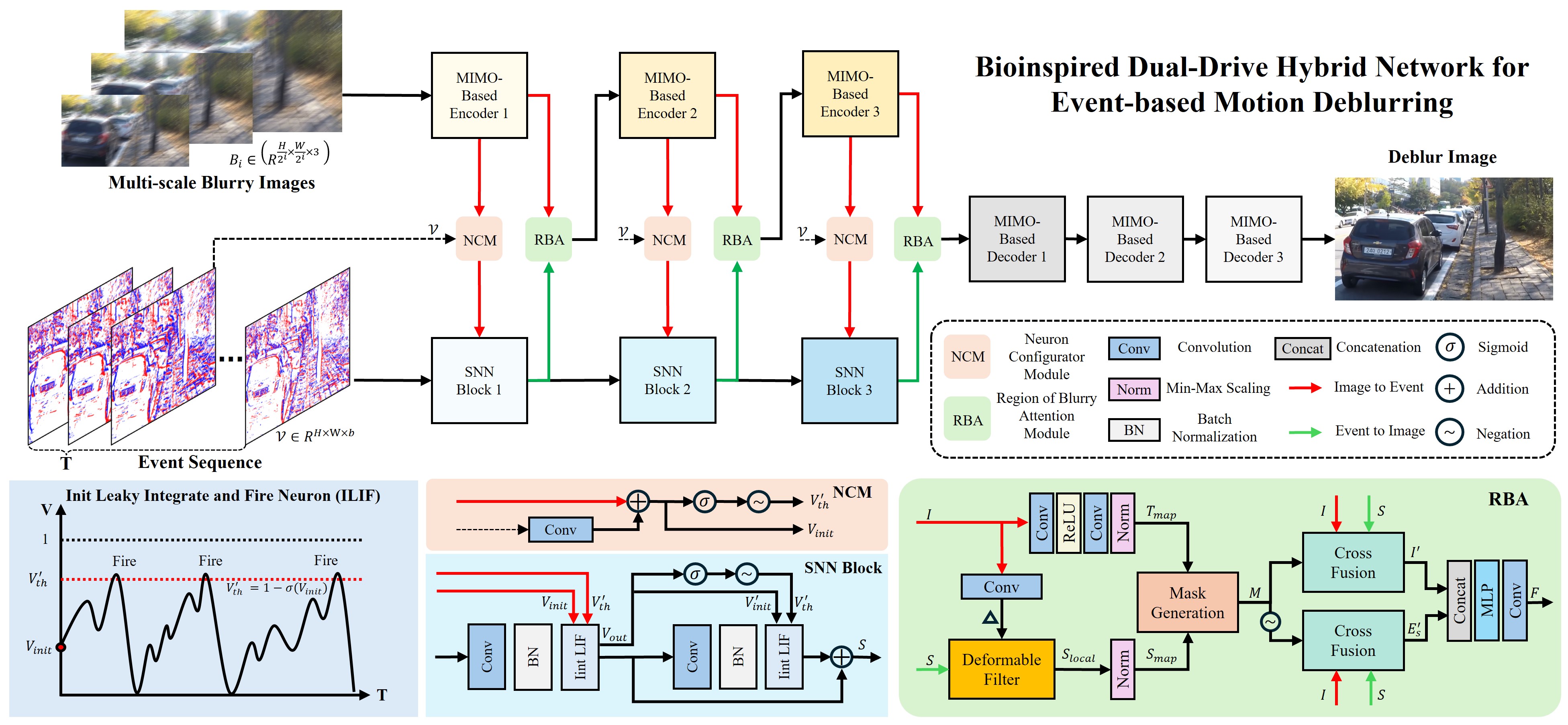}
   \caption{The overall framework of BDHNet. The event stream is shaped into the voxel-based representation $\mathcal{V}$. $B_{i}$ are the multi-scale blurry images. $I$ and $S$ denote the image and spike features respectively. $V_{init}$ and $V^{\prime}_{th}$ are the initialized membrane potential and threshold. $T_{map}$ and $S_{map}$ stand for the local spike map and threshold map for the blurry mask generation. $M$ is the region of blurry mask. $I^{\prime}$ and $E^{\prime}_{s}$ are the image and event features after cross attention. The modules of the same name are shaded darker to indicate deeper network layers.}
\label{fig:framework}
\end{center}
\end{figure*}
\section{Method}

\subsection{Problem Formulation}

The event-based motion deblurring network is informed by the human visual system, which efficiently manage complex environmental conditions. This capability is reflected in the network's design, where visual stimuli are methodically broken down and processed in a hierarchical and parallel manner as color and motion, as shown in Figure~\ref{fig:motivation}.

In the motion deblurring task, the conventional RGB camera captures the color and texture details of the scenario, and event camera provides the motion information. The blur accumulation process can be modeled by the intensity of sharp images $\mathcal{I}(t)$ as:
\begin{equation}
    \mathcal{B} = \frac{1}{T} \int_{f-T/2}^{f+T/2} \mathcal{I}(t) \, dt.  
\label{eq： blur process}
\end{equation}
where $\mathcal{B}$ denotes the blurry image, $f$ indicates the latent time stamp of the sharp image and $T$ is the exposure period of the sensors.

For the bioinspired event camera, events are emitted asynchronously each time the log-scale brightness change exceeds the positive event threshold $c > 0$:
\begin{equation}
    \log(\mathcal{I}(t, x)) - \log(\mathcal{I}(f, x)) = p \cdot c,  
\label{eq： event}
\end{equation}
where $\log\left(\mathcal{I}(t, x)\right)$ and $\log\left(\mathcal{I}(f, x)\right)$ denote the log-scale intensity of pixel $x$ at time $t$ and $f$, and $p$ is the polarity of event data. 

$\mathcal{I}(t)$ can be computed based on the events generated by the current pixel within the exposure period $\forall t \in T$ as:
\begin{equation}
    \mathcal{I}(t) = {\mathcal{I}(f)}  \cdot \exp\left( c \cdot \int_f^t p(s) \, ds \right),
\label{eq： sharp process}
\end{equation}
where $\mathcal{I}(f)$ is the latent sharp image. 

Substitute Equation \eqref{eq： sharp process} into Equation \eqref{eq： blur process}, we deduce the following formula:
\begin{equation}
\mathcal{I}(f) = \frac{\mathcal{B}}{\frac{1}{T}\int_{f-T/2}^{f+T/2}\exp\left( c \int_f^t p(s) \, ds \right) \, dt},
\label{eq：simple}
\end{equation}

Since the direct restoration of $\mathcal{I}(f)$ via Equation~\eqref{eq：simple} often faces challenges due to the instability of event threshold $c$, learning-based methods are employed to more accurately model the statistical characteristics of events $\mathcal{E}$ as:
\begin{equation}
\mathcal{I}(f) = \mathrm{Deblur}(f; \mathcal{B},\mathcal{E}), \quad \forall f \in T,
\label{eq：deblur}
\end{equation}
where $\mathrm{Deblur}(\cdot)$ indicates a motion deblurring network.

\subsection{Network Architecture}

The overall framework of our proposed Bioinspired Dual-Drive Hybrid Network is shown in Figure~\ref{fig:framework}. We
adopt a classical encoder-decoder architecture for our approach. Initially, the multi-scale
blurry images $\mathcal{B} \in (\mathbb{R}^{H \times W \times 3},\mathbb{R}^{\frac{H}{2} \times \frac{W}{2} \times 3},\mathbb{R}^{\frac{H}{4} \times \frac{W}{4} \times 3})$ are fed into the ANN-based image branch, where we use the MIMO-Based Encoder \cite{cho2021rethinking} as our fundamental block, to extract relevant features. Simultaneously, the corresponding event stream is shaped into the voxel-based representation $\mathcal{E} \in \mathbb{R}^{H \times W \times b}$ and fed into the SNN-based event branch, which facilitates the extraction of motion features. Following each layer of the two branches, the Dual-Drive Enhancement is employed to mimic the visual attention in the human visual system. It consists of a Neuron Configurator Module (NCM) for dynamic setting the neuron configurations and a Region of Blurry Attention Module (RBAM) that strategically focuses on motion features causing blurry effects to enhance cross-modal feature fusion. In the decoder, the MIMO-Based Decoder is applied for the image reconstruction and the PSNR Loss function \cite{chen2021hinet} is applied to precisely optimize the network’s parameters for optimal performance.

\subsubsection{Neuron Configurator Module}

The Neuron Configurator Module (NCM) is designed for visual enhancement from image data to event data. It strategically modulates neuronal responses to concentrate on critical regions as the neuron-based attention. Unlike previous approach \cite{aydin2024hybrid} that initialize the neuron's membrane potential at the first timestep, NCM employs image features to set both the initial timestep membrane potentials and the neuron thresholds across all timesteps based on input stimuli characteristics, enabling a pixel-level dynamic adjustment. Specifically, the structural and chromatic features, initially extracted from the blurry images $\mathcal{B}$, are utilized to identify the blurry regions. These features are then integrated with the attributes extracted from event data $\mathcal{E}$ through a shallow convolution layer, designed to mitigate domain discrepancies between the two modalities. 

The initial membrane potential ${V}_{\text{init}}$ at $t=0$ is set based on the integrated features from blurry image $\mathcal{B}$ and event data $\mathcal{E}$ as:
\begin{equation}
{V}_{\text{init}} = \phi_{\text{init}}(\mathcal{B}) + \psi_{\text{init}}(\mathcal{E}),
\label{eq：init}
\end{equation}
where $\phi_{\text{init}}$ is the feature encoder of the blurry image and $\psi_{\text{init}}$ is the shallow convolution layer of the event data.

\begin{figure*}[ht]
\begin{center}
\includegraphics[width=0.9\linewidth]{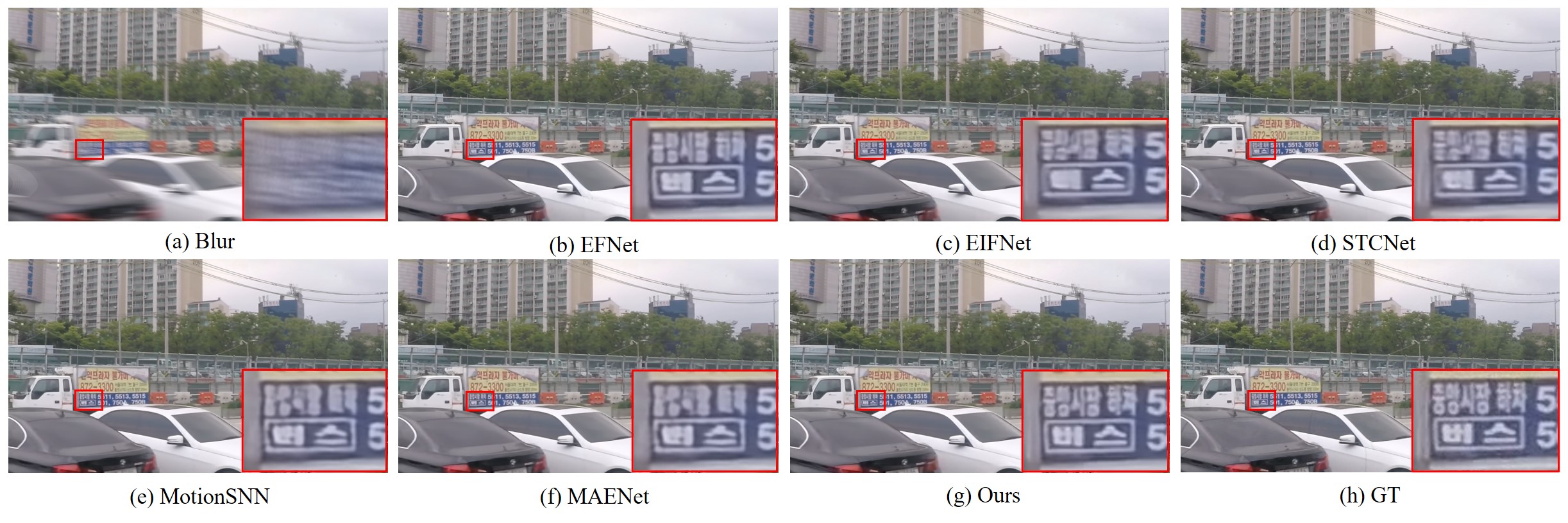}
   \caption{Qualitative comparisons under GoPro dataset. Best viewed on a screen and zoomed in.}
\label{fig:GOPRO}
\end{center}
\end{figure*}

To fully elevates neuron activity across blurry areas, the  threshold ${V}_{\text{th}}^{\prime}$ is redesigned according to the initial membrane potential as:
\begin{equation}
{V}_{\text{th}}^{\prime} = 1 - \sigma({V}_{\text{init}}),
\label{eq：threshold}
\end{equation}
where $\sigma$ is the Sigmoid function, which rescales the initial feature to the range of 0 to 1. Unlike the global scalar threshold $V_{\mathrm{th}} \in \mathbb{R}$ in vanilla LIF in Equation~\eqref{eq:preliminary:s=f(u-vth)} and \eqref{eq:preliminary:reset}, the threshold ${V}_{\text{th}}^{\prime}\in \mathbb{R}^{H \times W \times C}$ has the same dimension with the feature map, providing more fine-grained control over the reset and firing processes of the neurons in the same layer.

After the neuron configuration, the membrane potential update formula of each neuron is simplified to utilize ${V}_{\text{init}}$ to directly set the membrane potential at the initial timestep, with subsequent timesteps updated as follows:
\begin{equation}
\boldsymbol{u}[t] = \begin{cases} 
{V}_{\text{init}} & \text{if } t = 0 \\
(1 - \frac{1}{\tau}) \boldsymbol{u}[t-1] + \boldsymbol{c}[t] & \text{otherwise}
\end{cases}.
\label{eq：update}
\end{equation}
The condition for spike emission and the membrane potential reset strategy are as follows:
\begin{equation}
\boldsymbol{s}[t] = \begin{cases} 
1, & \text{if } \boldsymbol{u}[t] \geq {V}_{\text{th}}^{\prime} \\
0, & \text{otherwise}
\end{cases}.
\label{eq：spike}
\end{equation}
After the spike, the soft reset process is updated as follows:
\begin{equation}
\boldsymbol{u}^{l}[t] =
 \boldsymbol{u}^l[t] - {V}_{\text{th}}^{\prime} \odot \boldsymbol{s}^l[t]
\label{eq:method:reset}.
\end{equation}

In the event data feature extraction branch, we have implemented an SNN Block with residual connections as shown in Figure~\ref{fig:framework}. This block comprises two layers: the first layer's neurons are dynamically configured at the pixel level with the cross-modal initialization. The membrane potentials from the first layer subsequently inform the initialization and configuration of the second layer. Outputs from both layers are fused through a residual connection, culminating in the final output spikes $S$. This configuration, facilitated by the Neuron Configurator Module, enables precise pixel-level adjustments of neuronal activity, enhancing the neurons' responsiveness in blurry regions and thus improving the extraction of motion features from event data.

\subsubsection{Region of Blurry Attention Module}

The Region of Blurry Attention Module (RBAM) is proposed for visual enhancement from event data to image data as shown in Figure~\ref{fig:framework}. The RBAM module capitalizes on image features $I$ from the ANN branch and spike features $S$ from the SNN branch to generate a mask $M$ delineating blurry regions. This mask serves to capture accurate motion clues from event features, thereby refining the cross-modal feature fusion. 

Specifically, spikes activated by multiple pixels within the SNN branch originate from the same moving object in the scenario, with pertinent information about the motion contours embedded in the blurry image. Accordingly, the RBAM employs a deformable filter for localized aggregation of spike features. 

\begin{figure*}[ht]
\begin{center}
\includegraphics[width=0.9\linewidth]{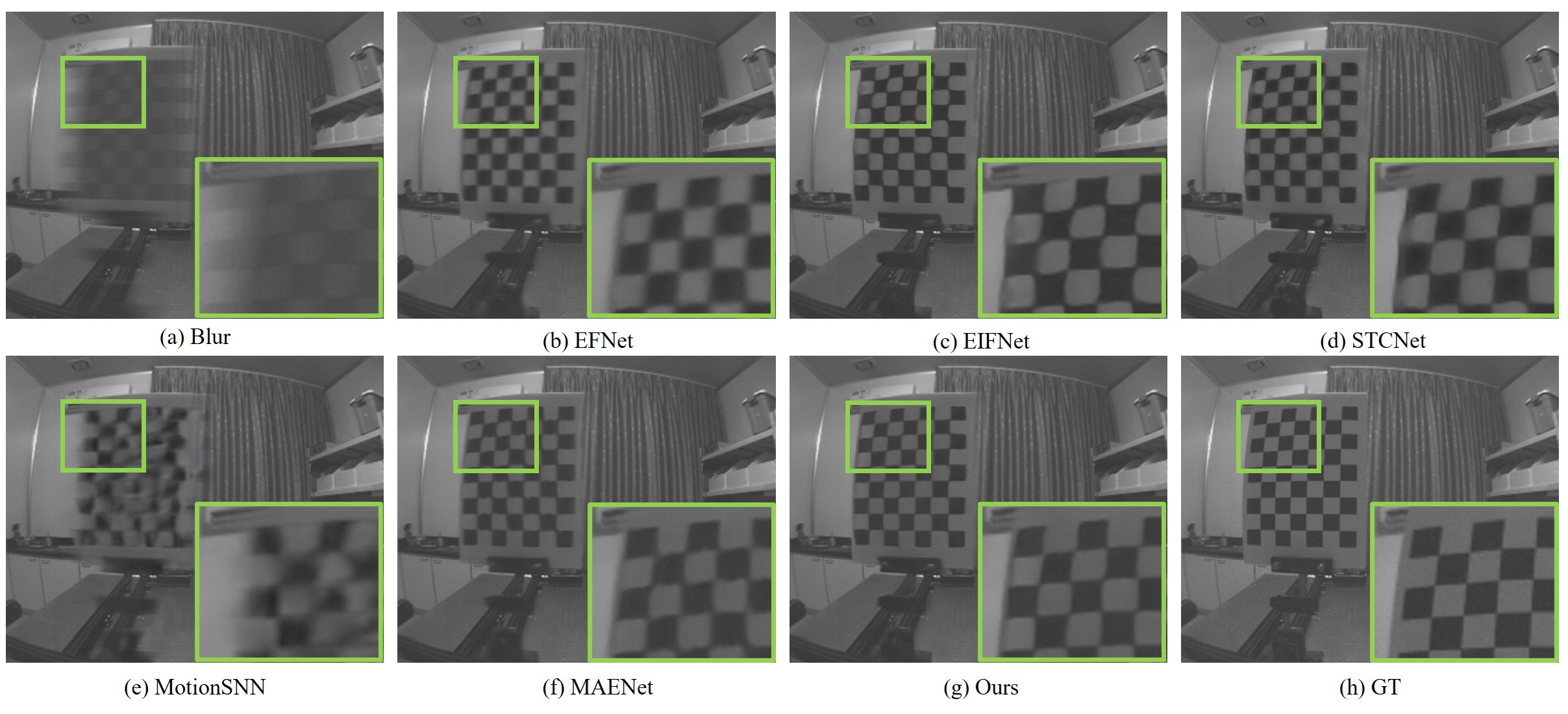}
   \caption{Qualitative comparisons under REBlur dataset. Best viewed on a screen and zoomed in.}
\label{fig:REBLUR}
\end{center}
\end{figure*}

\begin{figure*}[ht]
\begin{center}
\includegraphics[width=0.9\linewidth]{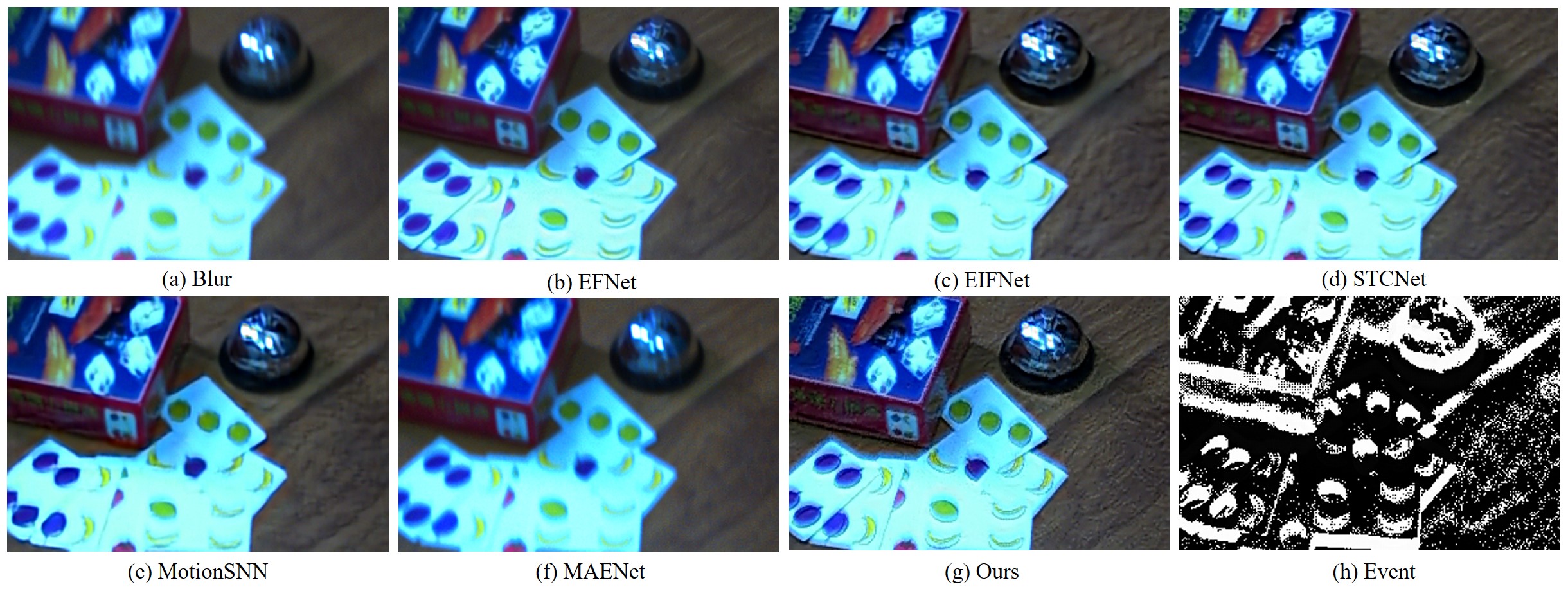}
   \caption{Qualitative comparisons under MS-RBD dataset. Best viewed on a screen and zoomed in.}
\label{fig:MS}
\end{center}
\end{figure*}

Within this filter, the spike features are firstly integrated in the temporal dimension as,
\begin{equation}
S_{sum} = \sum_{t} S(x, y, t).
\label{eq：sum}
\end{equation}
The bias for each deformable convolution kernel associated with individual pixels is determined by image features processed through a shallow convolution layer and the deformable filter is formulated as, 
\begin{equation}
S_{local} = \text{DeformSum}(S_{sum}, \text{Conv}(I)),
\label{eq：deform}
\end{equation}

Subsequent to the aggregation by the deformable filter, the spike map $S_{map}$ undergoes min-max normalization to facilitate uniformity as follows:
\begin{equation}
S_{map} = \text{Norm}(S_{local}),
\label{eq：norm}
\end{equation}
where $\text{Norm}$ is the Min-Max scaling operation as $\text{Norm}(X) = \frac{X - \min(X)}{\max(X) - \min(X)}$.

Further, a pixel-level threshold map $T_{map}$ is generated through the image features as,
\begin{equation}
T_{map} = \text{Norm}(\text{Conv}(\text{ReLU}(\text{Conv}(I)))),
\label{eq：tmap}
\end{equation}
this map is then subjected to pixel-level binarization against the spike map, producing a mask that identifies the blurry regions as follows,
\begin{equation}
M = \begin{cases} 
1 & \text{if } S_{map} \geq T_{map} \\
0 & \text{otherwise}
\end{cases} .
\label{eq：mask}
\end{equation}

This unsupervised identification process for blurry areas significantly enhances cross-modal feature fusion by utilizing this dynamically generated mask. 

For the mask-guided fusion, the spike features undergo a temporal convolution to adaptively integrate the temporal information and obtain the event feature $E_s$. The image feature $I$ and the event feature $E_s$ are applied with the cross attention operation as,
\begin{equation}
\begin{cases} 
I^{\prime} = I + \text{mask} \cdot \text{Attention}(Q_I, K_{E_s}, V_{E_s}), \\
{E_s}^{\prime} = E_s + (1 - \text{mask}) \cdot \text{Attention}(Q_{E_s}, K_{I}, V_{I}),
\end{cases}
\label{eq：mask}
\end{equation}
where $\text{Attention}$ denotes multi-head attention operation. 

The fusion features are generated through the channel-wise concatenation as,
\begin{equation}
F = \text{Conv}(\text{MLP}(\text{Concat}(I^{\prime}, {E_s}^{\prime}))).
\label{eq：fusion}
\end{equation}

The final deblur images are reconstructed through the MIMO-based Decoder \cite{cho2021rethinking} and the entire training process is conducted in an end-to-end fashion.

\begin{table*}[ht]
\centering
\caption{Performance comparison on GoPro and REBlur datasets with and without fine-tune. The best results are in bold.}
\scalebox{0.88}{
\begin{tabular}{ccccccccccc}
\toprule
\multirow{2}{*}{Method} & \multicolumn{2}{c}{Input} & \multicolumn{2}{c}{GoPro} & \multicolumn{2}{c}{REBlur} & \multicolumn{2}{c}{REBlur w/o Fine-tune} \\
\cmidrule(r){2-3} \cmidrule(r){4-5} \cmidrule(r){6-7} \cmidrule(r){8-9}
 & Image & Events & PSNR $\uparrow$ & SSIM $\uparrow$ & PSNR $\uparrow$ & SSIM $\uparrow$ & PSNR $\uparrow$ & SSIM $\uparrow$ \\
\midrule
SRN \cite{tao2018scale} & \checkmark & $\times$ & 30.26 & 0.934 & 35.10 & 0.961 & \textbackslash & \textbackslash \\
HINet \cite{chen2021hinet} & \checkmark & $\times$ & 32.71 & 0.959 & 35.58 & 0.965 & \textbackslash & \textbackslash \\
NAFNet \cite{chen2022simple} & \checkmark & $\times$ & 33.69 & 0.967 & 35.48 & 0.962 & \textbackslash & \textbackslash \\
Restormer \cite{zamir2022restormer} & \checkmark & $\times$ & 32.92 & 0.961 & 35.50 & 0.959 & \textbackslash & \textbackslash \\
MSDI-Net \cite{li2022learning} & \checkmark & $\times$ & 33.28 & 0.964 & 36.14 & 0.968 & \textbackslash & \textbackslash \\
UFPNet \cite{fang2023self} & \checkmark & $\times$ & 34.06 & 0.968 & 36.11 & 0.968 & \textbackslash & \textbackslash \\
EFNet \cite{sun2022event} & \checkmark & \checkmark & 35.46 & 0.972 & 38.02 & 0.975 & 27.43 & 0.898 \\
MotionSNN \cite{liu2023motion} & \checkmark & \checkmark & 35.18 & 0.971 & 36.32 & 0.968 & 34.63 & 0.957 \\
EIFNet \cite{yang2023event} & \checkmark & \checkmark & 35.99 & 0.973 & 37.81 & 0.976 & 35.75 & 0.965 \\
STCNet \cite{yang2024motion} & \checkmark & \checkmark & 36.45 & 0.975 & 37.78 & 0.976 & 35.64 & 0.966 \\
MAENet \cite{sun2025motion} & \checkmark & \checkmark & 36.07 & 0.976 & 38.46 & 0.978 & 32.86 & 0.950 \\
Ours & \checkmark & \checkmark & \textbf{37.04} & \textbf{0.977} & \textbf{38.50} & \textbf{0.978} & \textbf{36.01} & \textbf{0.967} \\
\bottomrule
\label{table1}
\end{tabular}}
\end{table*}

\section{Experiment}

\subsection{Datasets}

We evaluate the proposed method with GoPro, REBlur and MS-RBD datasets containing both synthetic and real-world scenarios. 

\textbf{GoPro}: We evaluate the deblurring performance on GoPro dataset \cite{sun2022event}, which is the benchmark dataset for the image motion deblurring. It consists of 3214 pairs of blurry and sharp images, with 2103 pairs for training and 1111 pairs for testing. The resolution of all images is $1280 \times 720$ and the blurry images are produced by averaging several adjacent high-speed sharp images. The event data is generated through the ESIM simulator. In this work, the raw event data is shaped into voxel-based representation for each image following EIFNet and the timestep in $\mathcal{V}$ is set to $b=12$.

\textbf{REBlur}: REBlur dataset \cite{sun2022event}, captured by DAVIS for real event-based motion deblurring, comprises 1,389 image pairs with 486 designated for training and 903 for testing. It contains diverse linear and nonlinear indoor motions. Each image has a resolution of 260×360, consisting of real-world event data along with the corresponding blurry and sharp images.

\textbf{MS-RBD}: MS-RBD dataset \cite{zhang2023generalizing} is the multi-scale blurry dataset captured in the real-world scenario. The dataset contains 32 sequences of data with 22 indoor and 10 outdoor scenes. The resolution of all images is $288 \times 192$ with the corresponding events. We evaluate the deblurring performance on MS-RBD with a focus on the generalization ability in the real-world scenes, where the blur caused by camera ego-motion and dynamic scenes.

\subsection{Implementation Details}

During the training process, we deploy our proposed network in the PyTorch framework on a single NVIDIA RTX 4090 GPU. The ADAM optimizer \cite{kingma2014adam} is utilized with an initial learning rate of $1 \times 10^{-4}$, which is scheduled to decrease at the 60th and 80th epochs, over a total of 120 epochs. For data augmentation, horizontal and vertical flipping, rotation, and random crop are applied. The crop size is set to 512 for the GoPro dataset. Fine-tuning on the REBlur dataset is conducted over 30 epochs with an initial learning rate of $1 \times 10^{-5}$. The crop size for REBlur is set to 256 and other configurations are kept the same as for GoPro. Our evaluation metrics include PSNR and SSIM.

\subsection{Comparison Experiments}

\begin{figure*}[ht]
\begin{center}
\includegraphics[width=0.98\linewidth]{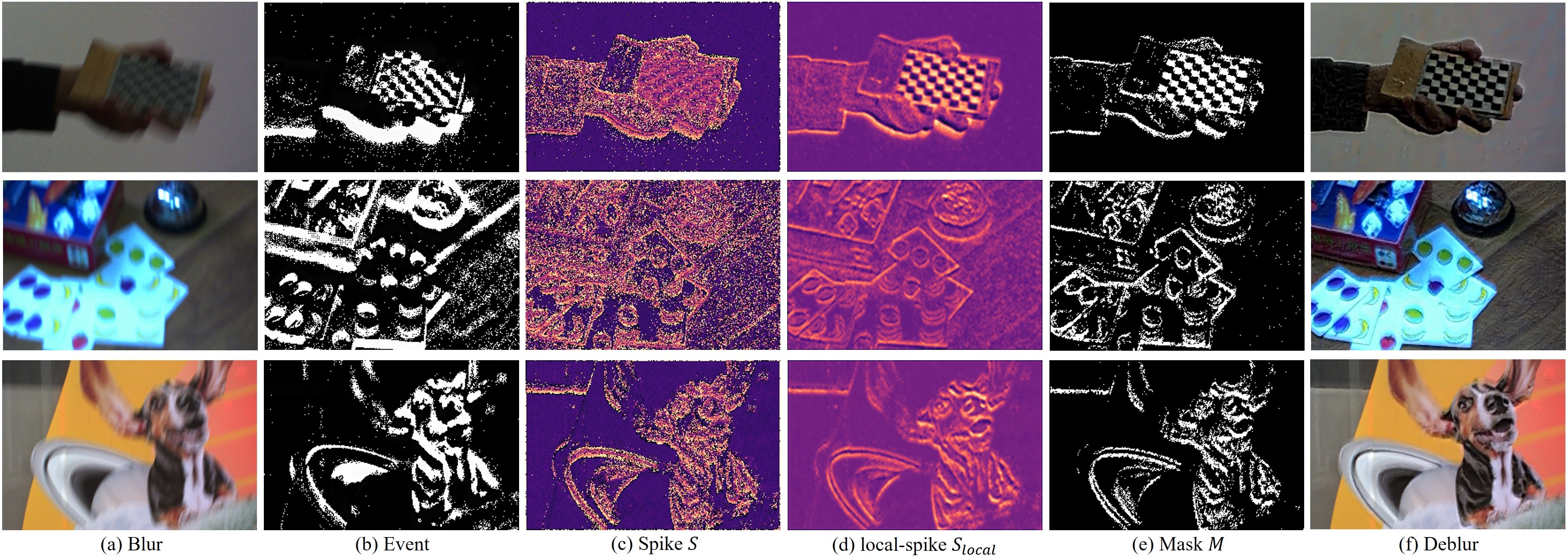}
   \caption{Visualization of the unsupervised blurry mask generation process under MS-RBD dataset.}
\label{fig:Abla_vis}
\end{center}
\end{figure*}

We compare our proposed BDHNet to SOTA image-only and event-based deblurring methods on GoPro, REBlur and MS-RBD datasets for a comprehensive evaluation. The comparison methods include image-only method: SRN \cite{tao2018scale}, HINet \cite{chen2021hinet}, NAFNet \cite{chen2022simple}, Restormer \cite{zamir2022restormer}, MSDI-Net \cite{li2022learning}, and UFPNet \cite{fang2023self}. The event-based methods consists of EFNet \cite{sun2022event}, MotionSNN \cite{liu2023motion}, EIFNet \cite{yang2023event}, STCNet \cite{yang2024motion}, and MAENet \cite{sun2025motion}. The event-based methods are all based on the raw event data produced by EFNet and we utilize the open-source checkpoint to evaluate the performance on GoPro dataset. For a fair comparison, all methods are trained under the optimal parameter settings as specified in the respective papers if there are no open-source checkpoint in REBlur dataset. Our comparison metrics follow the benchmark established by EFNet and MAENet, maintaining consistency in metric calculations libraries.

\begin{table}[t]
\centering
\caption{Ablation study of the proposed method on GoPro dataset. Image means input blurry image, Event stands for the corresponding event data, NCM is the Neuron Configurator Module, Mask is the region of blurry mask in RBAM, CA means cross attention, add is the addition operation. The best results are in bold.}
\scalebox{0.8}{
\begin{tabular}{ccccccc}
\toprule
Image & Event & NCM & Mask & Fusion Module & \multicolumn{1}{c}{PSNR / SSIM} \\
\midrule
\checkmark & $\times$ & $\times$ & $\times$ & $\times$ & 31.64 / 0.949 \\
\checkmark & \checkmark & $\times$ & $\times$ & Add & 36.29 / 0.972 \\
\checkmark & \checkmark & \checkmark & $\times$ & Add & 36.51 / 0.973 \\
\checkmark & \checkmark & \checkmark & $\times$ & CA & 36.60 / 0.974 \\
\checkmark & \checkmark & \checkmark & \checkmark & Add & 36.84 / 0.975 \\
\checkmark & \checkmark & \checkmark & \checkmark & CA & \textbf{37.04} / \textbf{0.977} \\
\bottomrule
\label{table_abla}
\end{tabular}}
\end{table}

Table~\ref{table1} provides a detailed comparative analysis of the comparison deblurring methods evaluated on the GoPro and REBlur datasets. Notably, our method significantly outperforms others in both datasets, achieving the highest performance metrics with a PSNR of 37.04 and an SSIM of 0.977 on the GoPro dataset, and a PSNR of 38.50 and an SSIM of 0.978 on the REBlur dataset. These results underscore our BDHNet's superior ability to mitigate blur effects under varied conditions.

Specifically, the performance on the REBlur dataset without fine-tuning is particularly noteworthy. Our method, when applied directly without specific adaptation to the REBlur dataset (trained solely on the GoPro data), achieves the best performance with a PSNR of 36.01 and an SSIM of 0.967. The robust generalization ability of our model derives from its bio-inspired architecture, which emulates the visual attention mechanism intrinsic to the human visual system. This design enables adaptive modulation of neuron responses, enhancing the model's focus on blurry regions. The neuron-based attention allows for effective adjustment to various blur intensities encountered across different datasets, obviating the need for dataset-specific tuning. This capability not only ensures consistent performance under diverse imaging conditions but also underscores the model's superiority for practical applications in real-world scenarios.

\begin{figure}[t]
\begin{center}
\includegraphics[width=0.9\linewidth]{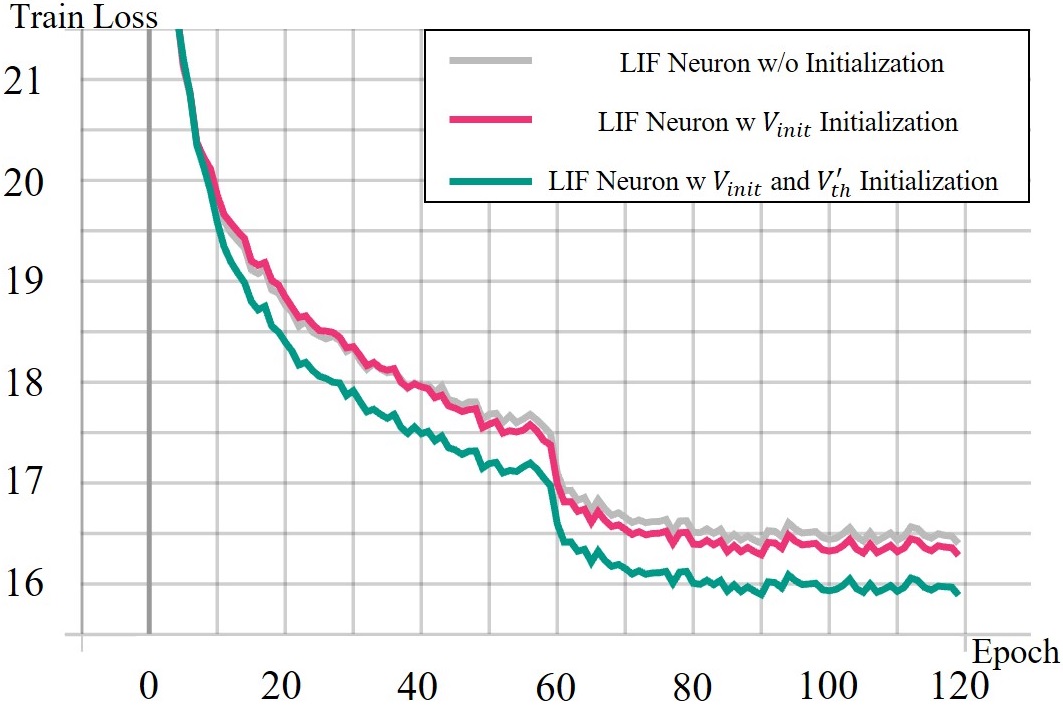}
   \caption{Training loss under different neuron configurations.}
\label{fig:neuron_loss}
\end{center}
\end{figure}

The visual comparisons presented in Figure~\ref{fig:GOPRO}, Figure~\ref{fig:REBLUR}, and Figure~\ref{fig:MS} effectively demonstrate the superior performance of our method in deblurring tasks, evidencing enhanced detail recovery and reduced spatial distortions across various scenarios. Our model consistently outperforms competing methods, achieving clearer and more precise reconstructions. Specifically, it excels in restoring sharper text in Figure~\ref{fig:GOPRO} and finer structural details in Figure~\ref{fig:REBLUR} and Figure~\ref{fig:MS}, significantly improving legibility and image quality. This showcases the effectiveness of our BDHNet in accurately perceiving and processing motion information, which is crucial for high-quality motion deblurring in practical applications. More visual comparison results are provided in the supplementary material.

The objective metrics and subjective evaluations in our study highlight our method's superior performance and exceptional generalization ability across diverse datasets. 

\subsection{Ablation Studies}

To evaluate the effectiveness of the key components in the proposed BDHNet, we conduct comprehensive ablation studies on the GoPro datasets, as shown in Table~\ref{table_abla}. We also visualize the generation process of the blurry region mask step by step in Figure~\ref{fig:Abla_vis}.

\begin{figure}[t]
\begin{center}
\includegraphics[width=0.98\linewidth]{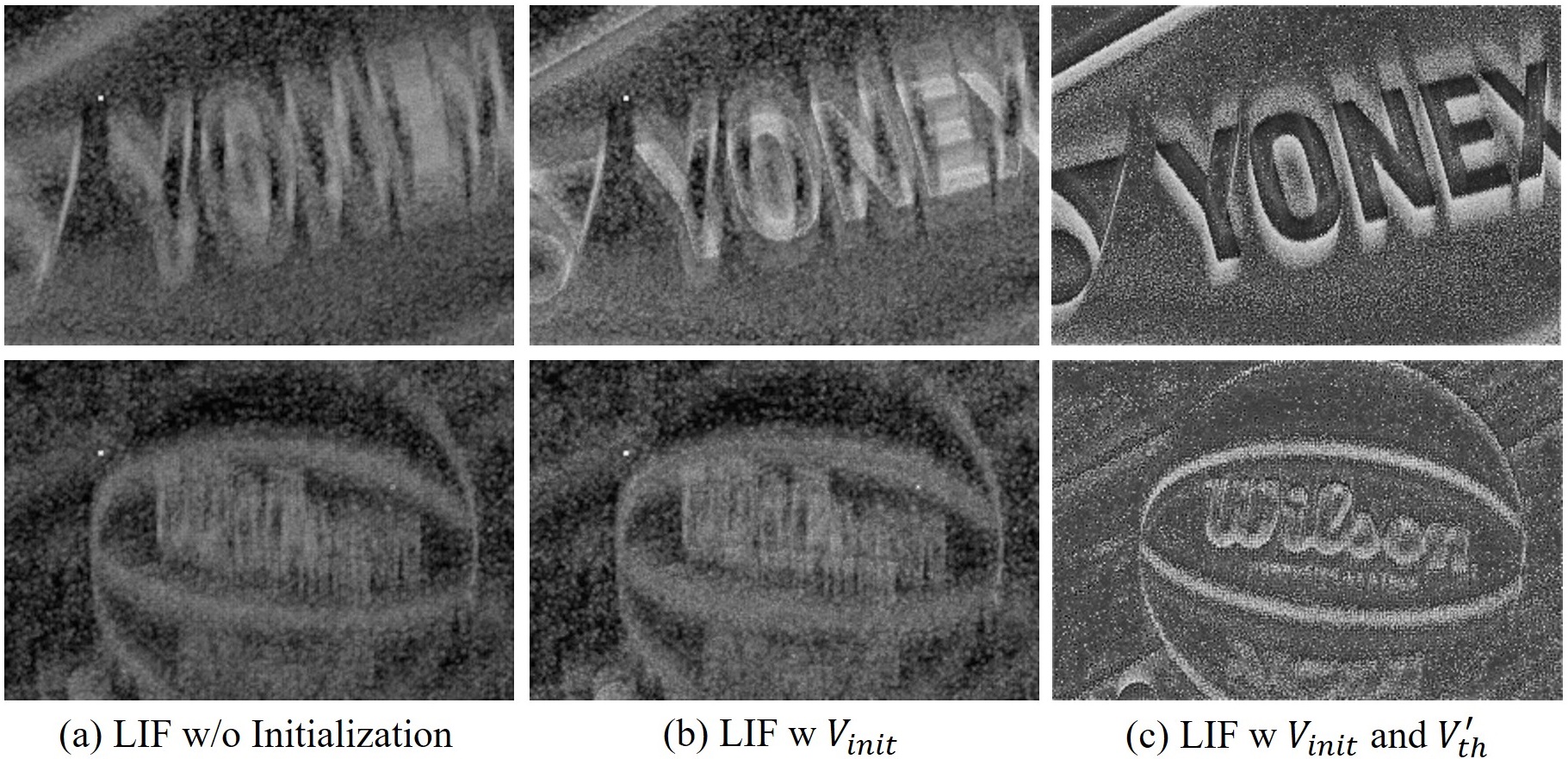}
   \caption{Visualization of the neuron responses of different configurations under MS-RBD dataset.}
\label{fig:neuron_vis}
\end{center}
\end{figure}

\textbf{Effectiveness of the Neuron Configurator Module.} The comparative analysis between the second and third rows of Table~\ref{table_abla} substantiates the efficacy of the Neuron Configurator Module as the neuron-based attention. Compared to the deblurring performance under conditions without initial neuron configuration, the PSNR increased by 0.22 dB. Our evaluation includes a focused comparison between methods that initialize only the membrane potential at the first timestep and our proposed approach, which encompasses the initialization of membrane potential and the dynamic configuration of neuron thresholds across all timesteps. As evidenced by the visual comparisons in Figure~\ref{fig:neuron_loss}, our method demonstrates superior training convergence. In contrast, initializing only the first timestep membrane potential \cite{aydin2024hybrid} offers marginal improvements under conditions without configuration. 

Figure~\ref{fig:neuron_vis} further validates the effectiveness of our configuration method by visualizing the spike outputs produced by neurons under three different settings for the same scene. It illustrates that our method, which leverages bio-inspired visual attention mechanisms from the human visual system, effectively modulates spikes triggered by varying visual stimuli through dynamic neuron configuration. This method proficiently concentrates spikes on blurry regions or motion-inducing edges. In contrast, configurations that either only initialize the initial membrane potential or without initialization struggle to capture motion characteristics effectively, thus yielding suboptimal deblurring results.

\textbf{Effectiveness of the Region of Blurry Attention Module.} In the RBAM module, our architecture is divided into two principal components. The first component focuses on generating a mask for blurry regions in an unsupervised manner that utilizes both spike and image features. The second component applies the mask to guide the cross-modal feature fusion. The efficacy of this masking process is substantiated by incremental improvements in PSNR of 0.33 dB and 0.44 dB, as shown in the comparative analysis between rows 3 and 5, and rows 4 and 6 in Table~\ref{table_abla}.

The mechanics of this unsupervised mask generation are detailed through the visualizations in Figure~\ref{fig:Abla_vis}, based entirely on the real-world MS-RBD dataset, thus demonstrating the robustness and the generalization ability of our approach. Column c and d of Figure~\ref{fig:Abla_vis} utilize heatmaps to visually demonstrate that spike features, post-processing with a deformable filter, are more accurately focused on the blurry areas or edges inducing blur, as illustrated in column d. This enhanced focus is facilitated by the deformable bias introduced by image features, which implicit captures motion information. The resulting masks, generated from the amalgamation of local aggregation spike maps and threshold maps derived from image features as depicted in column e, effectively delineate the regions of blurriness. The region of blurry mask accurately captures the motion clues that cause the blurry effects from the event features, thereby guiding the effective cross-modal feature fusion.

We further evaluate the effectiveness of the generated masks with various feature fusion methods: pixel-level addition and cross-modal attention. According to the data presented in row 5 of Table~\ref{table_abla}, our approach utilize addition for fusion exhibited significant deblurring capabilities. Subsequently, integrating a cross attention fusion strategy further augmented our model's performance, enabling it to reach state-of-the-art level in motion deblurring.



\section{Conclusion}

In this paper, we propose the Bio-inspired Dual-Drive Hybrid Network (BDHNet) for event-based motion deblurring. Drawing inspiration from the human visual system, the dual-drive enhancement strategy effectively mitigates the impact of blur resulting from camera or scene motion. The integration of the Neuron Configurator Module (NCM) and the Region of Blurry Attention Module (RBAM) enables dynamic and precise adaptation to blurry areas. Comprehensive evaluations demonstrate that BDHNet sets a new standard in the field, surpassing existing technologies in both synthetic and real-world scenarios. 


\end{document}